\pdfoutput=1

\documentclass[11pt]{article}

\usepackage[final]{acl}

\usepackage{times}
\usepackage{latexsym}

\usepackage[T1]{fontenc}

\usepackage[utf8]{inputenc}

\usepackage{microtype}

\usepackage{inconsolata}

\usepackage{graphicx}

\usepackage{tabularx}
\usepackage{amsmath}
\usepackage{mathtools}
\usepackage{amssymb}
\usepackage{algorithm}
\usepackage{algpseudocode}
\usepackage{cleveref}
\usepackage{multirow}
\usepackage[export]{adjustbox}
\usepackage{subcaption}
\usepackage{soul}
\usepackage{fontawesome}
\usepackage[shortlabels]{enumitem}
\usepackage{booktabs}
\usepackage{xspace}

\newsavebox{\largestimage}

\usepackage{listings}

\usepackage{xcolor}

\lstdefinestyle{promptstyle}{
    basicstyle=\small\ttfamily,
    breaklines=true,
    breakatwhitespace=true,
    frame=single,
    backgroundcolor=\color{gray!10},
    columns=fullflexible,
    keepspaces=true
}

\title{Breaking the Autoregressive Chain: Hyper-Parallel Decoding for Efficient LLM-Based Attribute Value Extraction}

\newcommand{\by}{\boldsymbol{y}}
\newcommand{\bx}{\boldsymbol{x}}
\newcommand{\ba}{\boldsymbol{a}}
\newcommand{\bv}{\boldsymbol{v}}
\newcommand{\bp}{\boldsymbol{p}}
\newcommand{\bs}{\boldsymbol{s}}

\newcommand{\hpd}{Hyper-Parallel Decoding\xspace}
\newcommand{\pluseq}{\mathrel{{+}{=}}}

\author{
 \textbf{Theodore Glavas\thanks{\hspace{1mm}This work was done as part of an internship at Amazon.}\textsuperscript{1,2,3,4}},
 \textbf{Nikhita Vedula\textsuperscript{1}},
 \textbf{Dushyanta Dhyani\textsuperscript{1}},
 \\
 \textbf{Yilun Zhu\thanks{\hspace{1mm}Work done while at Amazon. Currently at Apple.}\textsuperscript{1}},
 \textbf{Shervin Malmasi\textsuperscript{1}}
\\
 \textsuperscript{1}Amazon.com, Inc.,
 \textsuperscript{2}McGill University,
\\
 \textsuperscript{3}Mila - Quebec AI Institute,
 \textsuperscript{4}Int. Lab. Learning Systems
\\
 \small{\texttt{theodore.glavas@mail.mcgill.ca, yz565@georgetown.edu, \{veduln, dhyanidd, malmasi\}@amazon.com}}
}

\begin{document}
\maketitle

\begin{abstract}
Some text generation tasks, such as Attribute Value Extraction (AVE), require decoding multiple independent sequences from the same document context.
While standard autoregressive decoding is slow due to its sequential nature, the independence between output sequences offers an opportunity for parallelism. We present Hyper-Parallel Decoding, a novel decoding algorithm that accelerates offline decoding by leveraging both shared memory and computation across batches.
HPD enables out-of-order token generation through position ID manipulation, significantly improving efficiency. 
Experiments on AVE show that 
attribute-value pairs are conditionally independent, enabling us to parallelize value generation within each prompt. By further stacking multiple documents within a single prompt, we can decode in parallel up to 96 tokens per prompt.
HPD works with all LLMs, and reduces both inference costs and total inference time by up to 13.8X without compromising output quality, potentially saving hundreds of thousands of dollars on industry AVE tasks.
Although designed for attribute extraction, HPD makes no assumptions unique to the AVE domain and can in theory be applied to other scenarios with independent output structures.
\end{abstract}

\begin{figure}[ht!]
  \centering
  \includegraphics[width=0.9\columnwidth]{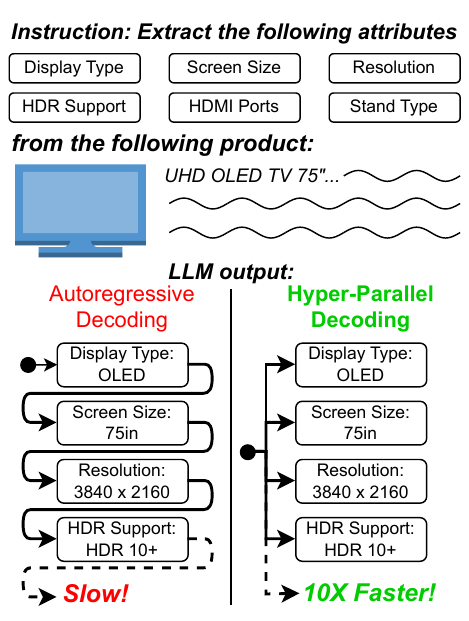}
  \vspace{-0.5em}
  \caption{AVE on the e-commerce domain: Given a set of attributes for the category \textit{Television}, values for each attribute are autoregressively generated using an LLM. Our method, \textbf{Hyper-Parallel Decoding (HPD)} parallelizes the extraction of values within the prompt to dramatically increase throughput and decrease cost.}
  \label{fig:figure1}
\end{figure}

\section{Introduction}
Information Extraction (IE) is a broad subcategory of Natural Language Processing (NLP) tasks that extract structured information from unstructured data. Generative IE has emerged from applying LLMs to IE in domains like e-commerce, medicine and real estate \citep{Brinkmann2024_ExtractGPT, agrawal2022_Medical_IE_LLM, Kvet2025_RAVE}, and recent work has shown that LLM-based IE is highly effective \citep{Roy2024_exploring_gen_frameworks}. However, the computational cost of generative LLMs limits their adoption in real-world scenarios \citep{zhang2025_GIE_survey}.
Prior work has proposed several techniques such as quantization, pruning and distillation \citep{zhou2024_efficient_llm_survey} to effectively accelerate LLM inference, but inference costs can still be prohibitive on large-scale data. 
Moreover, these approaches are fundamentally limited by the autoregressive decoding process, which inherently limits the potential for parallelization on efficient hardware such as GPUs. This motivates our central research question: \textbf{\textit{How can we overcome the autoregressive bottleneck to dramatically increase the speed of LLM-based IE and reduce cost, while maintaining output quality?}}

We focus on Attribute Value Extraction (AVE), a sub-task of IE.
AVE requires extracting values for multiple predefined attributes from a single document (see \Cref{fig:figure1}).
Since AVE often needs to be performed on massive sets of documents, it can greatly benefit from offline batched inference. 
Specifically, e-commerce is a popular domain for AVE as product listings often contain noisy, unstructured text. Crucial applications such as search \citep{lu-etal-2021-graph, xiao-etal-2021-end}, recommendation \citep{HWANGBO201894, truong2022ampsum} and question answering \citep{zhang-etal-2020-answerfact, rozen-etal-2021-answering} can greatly benefit from \textit{structured} information in the form of attribute-value pairs.

Autoregressive decoding is slow due to its sequential nature \citep{shazeer2019_fasttransformer}. However, for tasks like AVE, outputs are independent and can be decoded in parallel.
Prior literature has explored the idea of decomposing LLM generation into parallel components to accelerate decoding speed. For example, \citet{ning2024_skeletonofthought} show that independent reasoning chains can be decoded in parallel for problem solving. Although single-batch latency is reduced, such methods do not translate to the offline setting where batched inference is used, and where throughput and cost are the target metrics. Approaches such as Medusa \citep{Cai2024_medusa} support batched inference and can increase throughput by proposing a chain of multiple consecutive tokens in parallel. However, as these tokens are dependent on each other, throughput is limited by the maximum length of the proposals and the verification step required to ensure quality.

To address the above challenges, we introduce \textbf{Hyper-Parallel Decoding (HPD)}, a novel inference method that can increase the parallelization of generative AVE, unlocking extreme throughput gains up to 13X. Our key insight is that different generative tasks (e.g. attribute values) can be treated as \textit{conditionally independent} given a common document context. HPD leverages this inherent structure of AVE to \textit{break the autoregressive dependency} during generation, and simultaneously generate multiple attribute-value pairs. We further introduce an additional parallelization dimension by also extracting attributes of multiple documents within a shared prompt in parallel, analogous to how CPU hyper-threading can simultaneously complete multiple instructions. Our findings demonstrate that generating non-consecutive, independent tokens in parallel is far more efficient for increasing throughput than prior work. Although we evaluate our approach on e-commerce datasets, HPD makes no assumption exclusive to this setting and can in theory be applied to any domain where the conditional independence of the LLM output holds. We make our code available.\footnote{\href{https://github.com/networkslab/HPD}{https://github.com/networkslab/HPD}} To summarize, our main contributions are: 
\paragraph{[1]} We introduce Hyper-Parallel Decoding, a novel, high speed inference method that can increase the parallelization of attribute-value pair extraction to generate up to 96 tokens per prompt.
\paragraph{[2]} We demonstrate compatibility with a variety of other cost-reduction techniques including quantization, knowledge distillation and batched inference, enabling strong performance in the cost-constrained offline inference setting.
\paragraph{[3]}We demonstrate the effectiveness of HPD on three e-commerce AVE benchmark datasets, resulting in no quality drop and achieving upwards of 13X execution time reduction and 13X monetary cost reduction for select LLMs. 

\section{Related Work}

\paragraph{AVE}
Early work in AVE formulated the problem as a sequence labeling task, identifying spans of the input document as attribute values. Tagging was performed using specialized domain rules \citep{Zhang2009_ontology,Vandic2012_search} and later generalized using BiLSTM-CRF \citep{huang2015_BI-LSTM_CRF} and BERT-based \citep{Xu2019_SU_OpenTAG} architectures. AVE was later formulated as a question answering (QA) task, with AVEQA \citep{Wang2020_AVEQA} and MAVEQA \citep{Yang2022_MAVEQA} using BERT \citep{devlin-2019_bert} and ETC \citep{ainslie2020_etc} encoders. Both tagging and QA-based AVE rely on the existence of a comprehensive labeled training set, and struggle to generalize to unseen/implicit attributes and values.  The generative IE paradigm \citep{blume-etal-2023-generative, Roy2024_exploring_gen_frameworks} takes advantage of LLMs to offer superior performance in AVE tasks with few labels or high data sparsity, which is common in the e-commerce domain. Recently, ExtractGPT \citep{Brinkmann2024_ExtractGPT} has shown that generating attribute-value pairs with LLMs for product AVE is highly effective \citep{Brinkmann2024_using_LLMs_PAVE,shinzato-etal-2023-unified, sabeh2024_empirical_comparison}.

\paragraph{Parallel LLM Inference} 
We limit our review to works which increase LLM inference parallelism, as other efficient LLM methods are largely orthogonal to our work.
Research in LLM inference parallelization often aims to reduce \textit{latency} in an online setting, where requests are streamed one at a time. Skeleton-of-Thought \citep{ning2024_skeletonofthought} breaks down complex reasoning problems into a set of parallel sub-tasks, which can be simultaneously decoded via API calls to an LLM service. This takes advantage of the unused batch dimension in online inference, but does not translate to the offline setting where we assume GPU VRAM to already be saturated by batching independent prompts. APAR \citep{liu2024_APAR} partially addresses this issue using paged-attention to share part of the common key-value cache across batch elements. However, the attention operation must still be repeated for every parallel stream, unlike our method which shares both the memory and computation within a single batch element.

Another common paradigm is to draft a chain of multiple \textit{consecutive} tokens in parallel, usually followed by a verification step \citep{Fu2024_lookahead_decoding,Lin2025_BiTA, Cai2024_medusa}. In this scenario, the level of attainable parallelism is limited by the quality of the token chain, which degrades for longer chains. Speculative decoding \citep{leviathan2023_spec_decoding} is similarly limited. In practice, the ratio of generated tokens to inference steps ranges from 2 to 5, as erroneous tokens must either be rejected by the verification step or contribute to quality degradation.

\begin{figure*}[ht]
  \centering
  \savebox{\largestimage}{\includegraphics[width=0.67\textwidth]{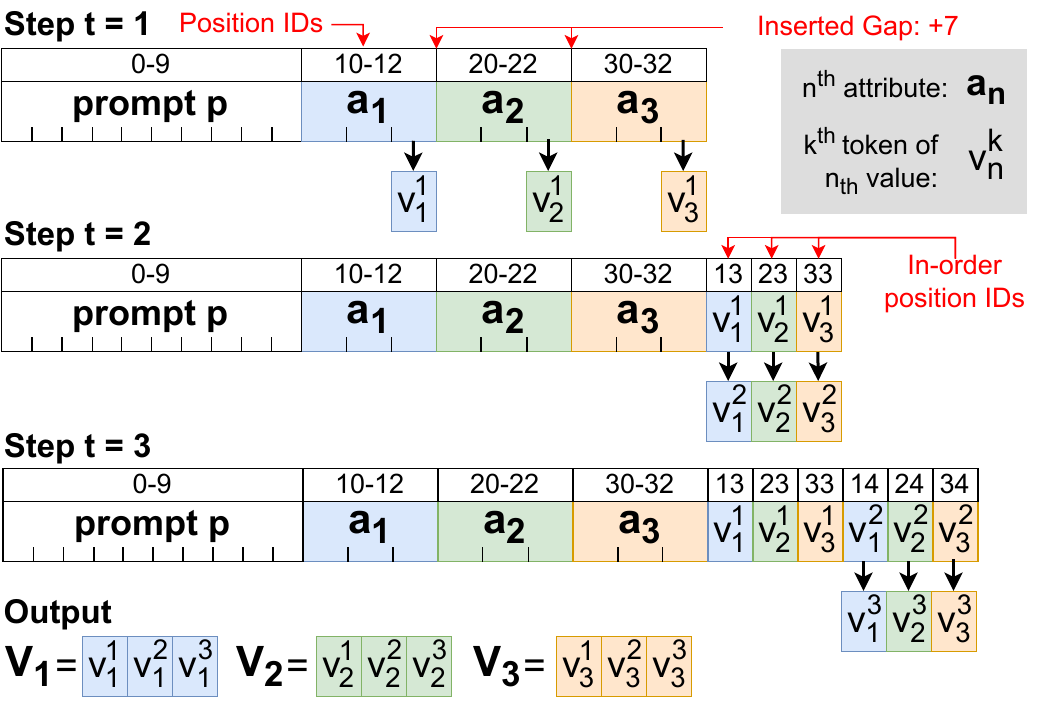}}%
  \begin{subfigure}{.32\textwidth}
    \centering
    \raisebox{\dimexpr.5\ht\largestimage-.5\height}{%
      \includegraphics[width=\linewidth]{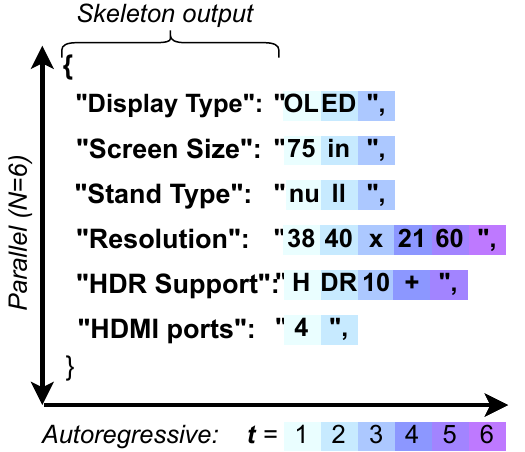}}
      \caption{ }
  \end{subfigure}
   \hfill
  \begin{subfigure}{0.67\textwidth}
      \centering
      \usebox{\largestimage}
      \caption{ }
  \end{subfigure}
  \caption{(a) An illustration of the generated output with HPD. The skeleton output defines the attributes to extract. The tokens within each value are generated autoregressively from left to right, while independent values are generated in parallel. (b) A block diagram for the HPD process: At step $t=1$, the first token of each value is decoded in parallel using the next token prediction at each attribute end position. The position IDs shown above each token have an inserted gap between attributes, which is gradually filled by the generated value tokens at each step. New tokens are always appended to the sequence in memory which places them out of logical order, but the position IDs define the attention mask and positional embeddings during attention. In this example, three values of three tokens are decoded in parallel using only three inference steps. An alternative version is presented in \Cref{fig:app_Figure2}.}
  \label{fig:HPD_flowchart}
\end{figure*}

\section{Attribute-Value Extraction (AVE) Task} \label{sec:problem_setting}
Our input is a set of documents $\bx_i \in \mathcal{X}$,
where $\bx_i = \{x_1,x_2,\dots\}$ is represented as a sequence of tokens in the vocabulary space $\mathcal{Y}$.
The documents are partitioned into categories $C(\bx_i) = c \in \mathcal{C}$. Each category $c$ has a list of predefined attributes $\ba_{1:N} = \{\ba_1,\ba_2,\dots,\ba_N\}$, where each $\ba_n \in \mathcal{A}$, $\ba_n = \{a_1,a_2,\dots\}$ is also represented as a sequence of vocabulary tokens. Note that we drop the indexing in $c$ to simplify the notation. The goal of AVE is to extract attribute values $\bv_n \in \mathcal{V}$, $\bv_n = \{v_1,v_2,\dots\}$ associated with each attribute for all products in $\mathcal{X}$. \Cref{fig:figure1} illustrates the AVE task for a product in the category \textit{Television}. 

In a generative LLM setting, we first construct a prompt $\bp(\bx_i,\ba_{1:N})$, as shown in \Cref{fig:HPD_flowchart}. The prompt contains instructions for the LLM, the attributes to extract and the input context. For each prompt, we wish to generate an output $\boldsymbol{y} = \{(\ba_1,\bv_1),(\ba_2,\bv_2),\dots,(\ba_N,\bv_N)\}$ containing the attribute/value pairs extracted from the input data.
In practice, $\by$ is a sequence of tokens $\{y_{1},y_{2},\dots\}$ consisting of the concatenated attributes/values in a structured text format.
We use JSON to represent the text (\Cref{fig:HPD_flowchart} (a)), but the approach is format-agnostic.
The full prompt appears in Appendix \ref{app:example_prompt}.

During inference step $t\in \{1,2,
\dots\}$, we provide the LLM a sequence consisting of the prompt $\bp(\cdot)$ concatenated with the partial output $\by_{1:t-1}$ if $t>1$, which we shorten to $\by_{<t}$. We define our decoder LLM network $F(\cdot)$ to output raw logit values over the vocabulary space $\mathcal{Y}$. Using autoregressive generation, we sample the next token from the random variable $Y_{t} \in \mathcal{Y}$. The conditional PDF of $Y_{t}$ is defined as the predicted probability distribution 
\begin{equation} \label{eq:LLM_predicted_P}
\hat{P}(Y_{t}=y \mid \bp, \by_{<t},) = \sigma( F(\by_{<t},\bp)) \in [0,1]^{\left|\mathcal{Y}\right|},
\end{equation}
where $\sigma(\cdot)$ is the softmax function.
Upon completion of the generation process after step $t=T$, we parse the generated tokens $\by_{1:T}$ to obtain the attribute-value pairs.

\paragraph{Metrics}
We evaluate AVE on the below metrics:
\begin{enumerate}[(i)]
    \item \textbf{Throughput:} 
    rate of products processed (product/s) on average per GPU with a benchmark server using wall-clock time. 
    \item \textbf{Cost:} GPU rental cost per 1K products processed (\$ / 1k products), or API cost for proprietary LLMs.
    \item \textbf{Output Quality:} F1 score of values against ground truth or pseudo-ground truth labels.
\end{enumerate}
 Specific implementation details are provided in Section \ref{sec:exp_setup}. Our overarching objective is to maximize the throughput and minimize the cost, while maintaining state-of-the-art output quality.

\section{Methodology}
We first describe the standard autoregressive decoding process from a probabilistic view in Section \ref{sec:autoregressive_gen}, and then describe our modifications made for HPD in Section \ref{sec:breaking_the_chain}. We follow with a detailed breakdown of the HPD algorithm in Section \ref{sec:HPD_method}.

\subsection{Autoregressive Generation} \label{sec:autoregressive_gen} In general LLM decoding, the Transformer architecture requires output tokens to be generated one at a time, i.e. \textit{autoregressively}. 
This constraint is imposed by the attention blocks.
At each step $t$, the current hidden state representation interacts with the attention keys and attention values derived from all previous token hidden states in the sequence. Since this hidden state is ultimately used to sample $Y_{t}$, there is a dependency for each $Y_{t}$ on all $Y_{<t}$.

For AVE, let us view the output $\by$ in the the attribute/value pair representation, $\by = \{(\ba_1,\bv_1),(\ba_2,\bv_2),\dots,(\ba_N,\bv_N)\}$. Let us denote $\boldsymbol{A}_n$ and $\boldsymbol{V_n}$ as the multivariate random variables representing the joint token distribution of the $n$-th attribute and value respectively. We can then express the full output random variable as a concatenation of the attribute and value random variables
\begin{align} 
    \boldsymbol{Y}_{n} &=concat(\boldsymbol{A}_n, \boldsymbol{V}_n) \nonumber\\ 
    \boldsymbol{Y}_{1:N} &=concat(\boldsymbol{Y}_{1},\boldsymbol{Y}_{2},\dots,\boldsymbol{Y}_{N}).
\end{align}
Since the attributes $\ba_n$ are known and provided in the prompt, we assume that their according random variable collapses to $P(\boldsymbol{A}_n = \ba_n) = 1$ for the correct attribute. We therefore omit $\boldsymbol{A}_n$ from subsequent equations and replace it directly with $\ba_n$. Since $\by$ is generated autoregressively, there exists a dependence between values:
\begin{align} \label{eq:value_autoregressive} 
    P(\boldsymbol{Y}_{1:N} \mid \bp)&=\prod_{n=1}^{N} P(\boldsymbol{V}_{n}\mid \bp, \ba_{1:n}, \boldsymbol{V}_{<n}).\\ \nonumber
    &=\prod_{n=1}^{N} P(\boldsymbol{V}_{n}\mid \bp, \boldsymbol{V}_{<n}).
\end{align}
The dependence on the attributes $\ba_{1:n}$ is omitted as it is already included in the prompt $\bp$. Equation \ref{eq:value_autoregressive}  shows that the $n^{th}$ value is dependent through attention on the previous $n-1$ values.

\subsection{Breaking the Autoregressive Chain}
\label{sec:breaking_the_chain}

For the AVE task, we argue that this value dependence is a negative consequence of autoregressive generation. The task is to extract values directly from the document presented in the prompt $\bp$. In theory, the extracted value for one attribute should not bias subsequent values. However, \citet{shinzato-etal-2023-unified} find that the ordering of attributes in the prompt can affect performance, as errors made in extracting the earlier values can cascade into errors and hallucinations in subsequent ones. 

We propose eliminating the inter-value dependence by breaking the fundamental inter-token dependence that underpins autoregressive generation. We assume \textbf{conditional independence} of generated values given the prompt $\bp$: 
\begin{align} \label{eq:value_independent}
    &P(\boldsymbol{Y}_{1:N} \mid \bp)=\prod_{n=1}^{N} P(\boldsymbol{V}_{n}\mid \bp).
\end{align}
Equation \ref{eq:value_independent} describes our assumption that the probability of generating value $\bv_{n}$ should not depend on previously generated values $\bv_{<n}$. By breaking the chain dependence, the predicted probability distribution of each predicted value
\begin{align}
    \hat{P}(\boldsymbol{V}_{n} = \bv_{n,1:K_{n}} \mid \bp) = \prod_{k=1}^{K_{n}}\sigma(F(\bv_{n,<k},\bp))
\end{align}
is independent from other values, although the $K_n$ tokens that the value $\bv_n$ comprises of must still be autoregressively generated. 
We can therefore compute $\hat{P}(\boldsymbol{V}_{n} = \bv)$ in parallel for all values, increasing the parallelism of the system. As a result, the number of LLM steps required for inference reduces from $T $ to $max(K_n) \equiv K_{max}$.

By breaking the autoregressive generation chain into $N$ smaller parallel chains for each value, we can express the predicted PDF of our output as:
\begin{align} \label{eq:final_HPD_equation}
\hat{P}(\boldsymbol{Y}_{1:N} = \by \mid \bp) &=\prod_{n=1}^{N} \hat{P}(\boldsymbol{V}_{n} \mid \bp)\\
    &= \underbrace{\prod_{n=1}^{N}}_{parallel}\underbrace{\prod_{k=1}^{K_{n}}\sigma(F(\bv_{n,<k},\bp)).}_{autoregressive} \nonumber
\end{align}

Next, we describe the algorithmic implementation of this process inside an LLM.

\subsection{Hyper-Parallel Decoding (HPD)}
\label{sec:HPD_method}

\subsubsection{Input and Position ID Construction} \label{sec:input_and_pos_id}
Given the assumed conditional independence of values in theory, we wish to parallelize the decoding of values using a standard decoder-only LLM.
\Cref{fig:HPD_flowchart} illustrates the basic \hpd process, which we refer to throughout this section. We begin by constructing the model input $\bs$ by concatenating the prompt $\bp(\bx,\ba_{1:N})$ with a skeleton output template of the desired output. An example skeleton output is shown as the uncolored text in Figure \ref{fig:HPD_flowchart} (a). In this template, we provide the attributes we wish to extract as keys, and leave the value field blank for the model to complete. For ease of notation, we simplify the output template to $\bs = cat(\bp,\ba_{1:N})$, ignoring the structure tokens like tabs, ``\{'' and ``\}''.
A block representation of $\bs$ is illustrated in the first row of Figure \ref{fig:HPD_flowchart} (b). In order to insert the value tokens into their respective positions during generation, a gap must be created in the output template.
We introduce this gap by \textbf{skipping position IDs}.
Position IDs are a variable used to define the absolute position of each token in a sequence, for example $\{0,1,2,3,\dots\}$.
LLMs are completely dependent on position IDs to encode the ordering of the input tokens, as they are used to define positional embeddings. At each position in $\bs$ where a value is to be generated, we increment the position IDs by $K_{max}$, thus leaving a gap in the position ID sequence. This hyperparameter defines the maximum length of values that can be generated. We provide the exact position ID calculation algorithm in Appendix \ref{app:position_id}.
In \Cref{fig:HPD_flowchart} (b), we insert a gap of $K_{max} = 7$ in the position IDs shown above each token block.
This manipulation allocates space for the values to be generated without using any additional memory.

\subsubsection{First Inference Step}
\begin{figure*}[ht]
  \centering
  \savebox{\largestimage}{\includegraphics[width=0.53\textwidth]{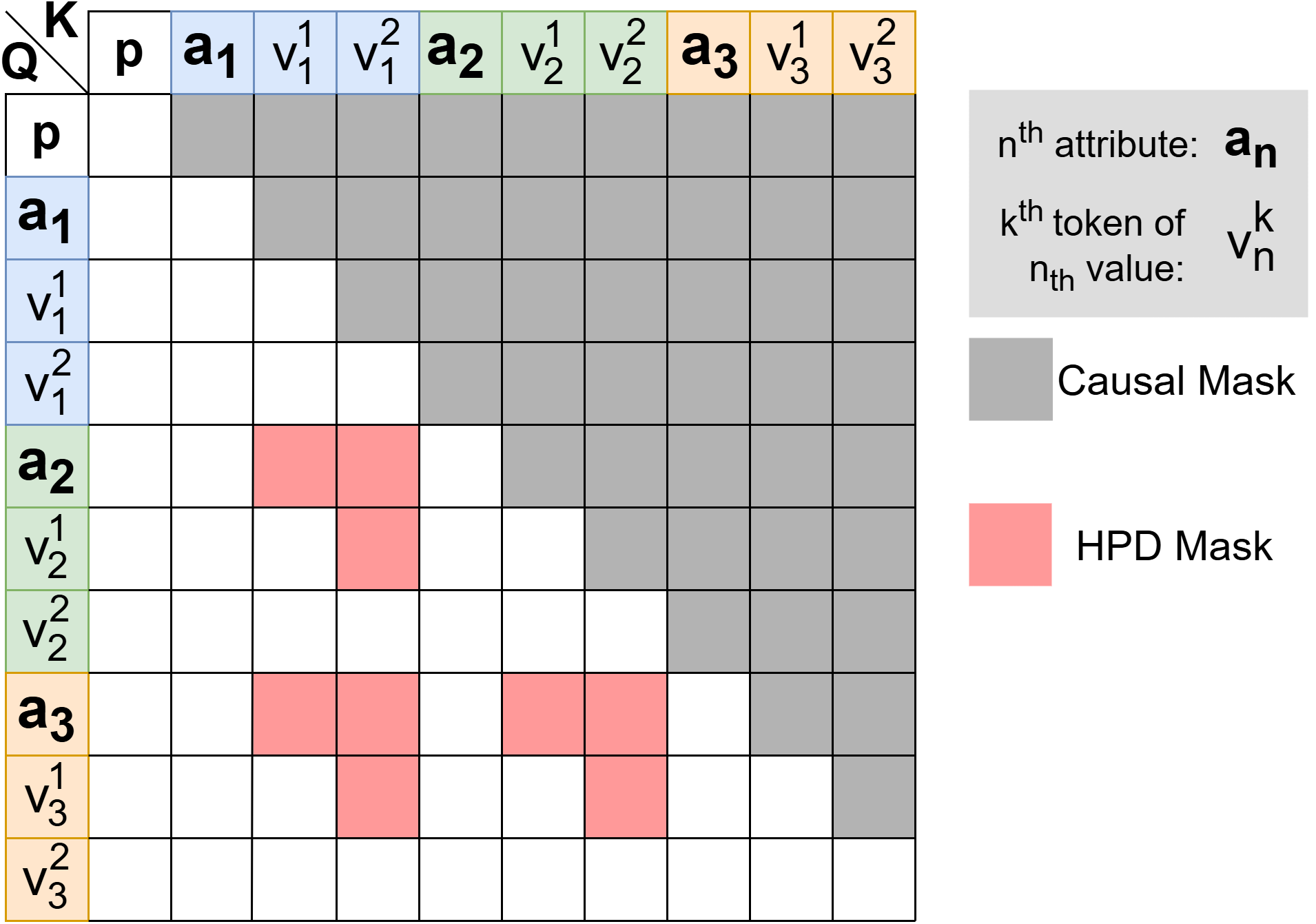}}%
  \begin{subfigure}{.37\textwidth}
    \centering
    \raisebox{\dimexpr.5\ht\largestimage-.5\height}{%
      \includegraphics[width=\linewidth]{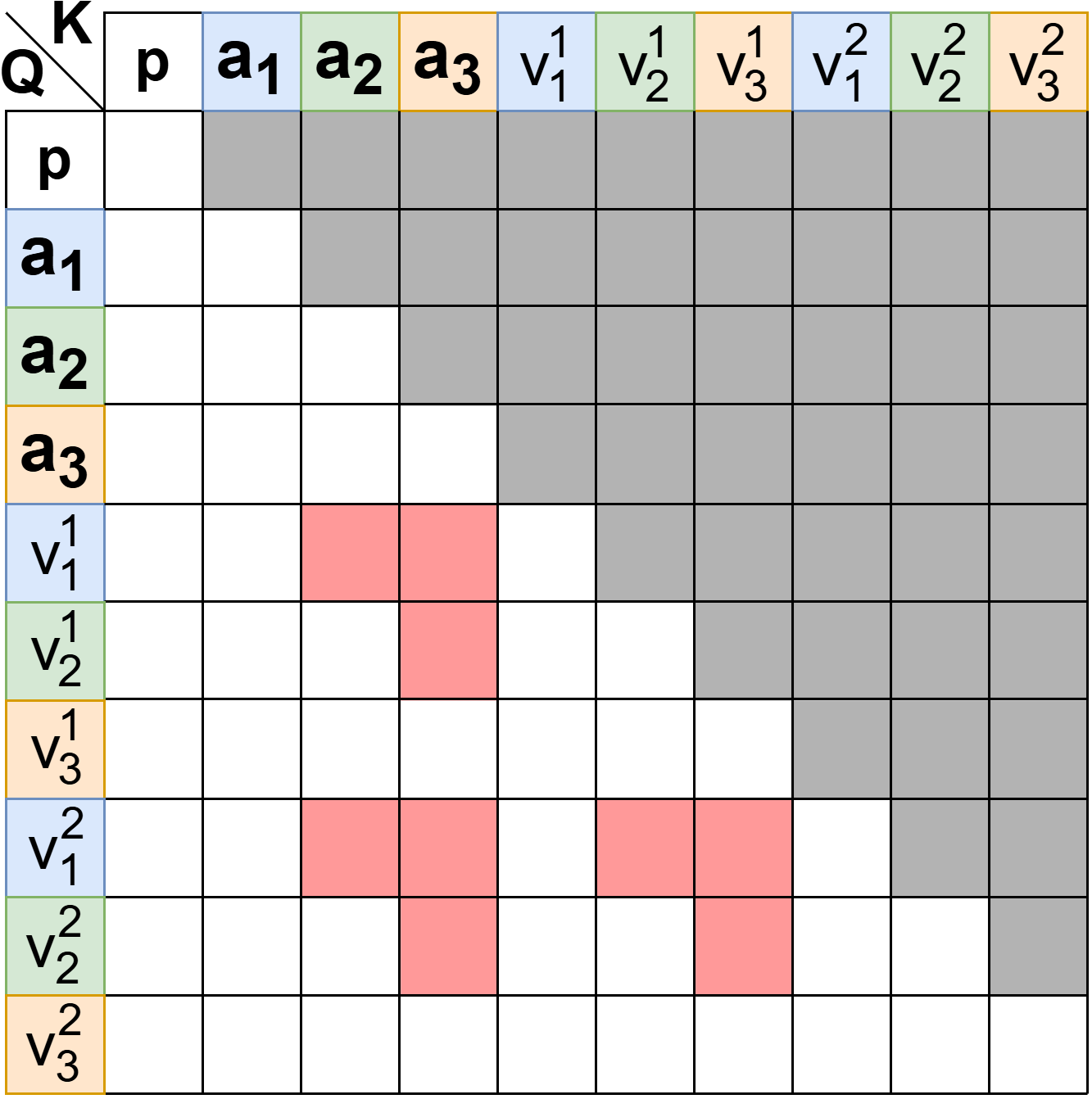}}
      \caption{Inference }
  \end{subfigure}
   \hfill
  \begin{subfigure}{0.53\textwidth}
      \centering
      \usebox{\largestimage}
      \caption{Fine-tuning }
  \end{subfigure}
  \caption{Illustration of the attention mask between queries Q (rows) and keys K (columns). Grey boxes indicate standard causal masking, while red boxes indicate the additional HPD mask applied. (a) During inference, decoded value tokens $v^k_n$ are placed out of order in memory relative to their position IDs. The HPD mask ensures causality by blocking attention to attribute and value tokens that appear later in the position ID ordered sequence. (b) The LLM is fine-tuned using ground truth or pseudo ground truth labels in the regular token order. The HPD mask addresses the train/test mismatch by masking attention between tokens which cannot attend to each other during inference due to the parallel decoding process. The attention pattern between (a) and (b) is identical.}
  \label{fig:HPD_mask}
\end{figure*}

During step $t=1$, the LLM outputs the next-token probability distribution $\sigma(F(\bs))$ for every token present in the input. Typically, only the last prediction is used, as it corresponds to the token probability for the next unknown token. We instead select the $N$ next-token probabilities at the position where the missing values are, and sample $N$ new tokens. These tokens make up the first token for each value $\{v_{1,1},v_{2,1},\dots,v_{N,1}\}$.
\Cref{fig:HPD_flowchart} (b) shows $N=3$ tokens decoded from positions 12, 22 and 32. 
Because the ordering of tokens is defined by their position IDs, we can append the new tokens to the end of the sequence, and assign them the position IDs that would place them in the gaps created in \Cref{sec:input_and_pos_id}. Appending new tokens to the end of the sequence is important for maintaining key-value cache functionality, since it is stored in a contiguous block of memory and cannot have values inserted mid-cache. In step $t=2$ of \Cref{fig:HPD_flowchart} (b), the first value tokens are given position IDs 13, 23 and 33, which would place them right after each of their respective attributes when ordering by position IDs. To maintain causal attention masking, the triangular attention mask is computed using the \textit{position ID ordering} of the tokens rather than their in-memory ordering. Using scaled dot-product attention, the use of custom attention masks is supported. \Cref{fig:HPD_mask} (a) illustrates how the attention mask is modified with HPD, using single tokens to represent the prompt and attributes. The resulting mask is non-triangular but enforces causal attention in the position ID ordering.

Although the \textit{ordering} is maintained, it is important to note that the values of the positions IDs are not, due to the existence of gaps in the sequence. We address this phenomenon later in \Cref{sec:HPD_fine_tuning}.

\subsubsection{Subsequent Inference Steps}
In subsequent inference steps $t>1$, only the $N$ new tokens are passed to the model, along with the previously computed key-value cache. The LLM can therefore generate the $t^{th}$ token for all $N$ values.
\Cref{fig:HPD_flowchart} (a) illustrates the order of decoding: tokens are generated autoregressively from left to right, but in parallel along the y-axis.
Position IDs are updated by incrementing the IDs used in the previous step, and the causal attention mask is computed with respect to these new positions.

\subsubsection{Early Stopping}
Since the length of individual values vary, we add a stop condition check at each inference step. For JSON decoding, we examine the generated token and prune values from the input sequence when the `$\backslash$n' character is detected.\footnote{Other structured formats require a different delimiter.} This pruning ensures that the key-value cache is not filled with irrelevant tokens.
Generation ends either when every value is pruned or after $K_{max}$ steps, where incomplete values are truncated.

\subsubsection{Parallelism with Document Stacking}
The current definition of the system generates at most $N$ tokens in parallel per inference step, which is limited by the number of attributes we wish to extract. We propose adding an additional parallelism dimension by decoding the values from multiple documents within a single prompt: we construct $\bp(\bx_{1:J},\ba_{1:N})$ by stacking $J$ documents %
in the same category. The instruction and attribute definition is efficiently shared among all documents, and the output of each document is stacked sequentially. Since the values of different products are also independent, we can now decode $J\times N$ tokens in parallel per inference step.
Prior work has referred to this technique as \textit{batch prompting} \citep{cheng2023_batch_prompting}, in the context of accelerating API calls. The speedup dynamics in this setting are very different, since the number of tokens decoded per step increases with $J$. Additionally, we do not ignore the associated performance penalty due to the longer prompt and larger key-value cache size, which reduces the maximum true batch size. In practice, we observe that significant gains can be attained from using $J>1$, but that there exists a point of diminishing returns as we show in \Cref{fig:HPD_scaling}. Appendix \ref{app:Algorithm} illustrates the entire HPD algorithm with document stacking.

\subsubsection{Support for Batch Inference}
A crucial benefit of HPD is the support for batching multiple prompts through the LLM at once. All of the modifications above described happen within a single prompt, meaning that multiple prompts can still be processed in parallel with standard batched inference.  When the number of incomplete values in each prompt becomes unbalanced, the length of the input tokens across batch elements is no longer equal, which is a requirement for combining batched inputs in tensor format. Therefore, the inputs are right padded with dummy tokens to maintain equal input lengths, and the dummy tokens are fully masked during attention. Combining all forms of parallelism, HPD can generate $b\times J\times N$ tokens per inference step for a batch size $b$, $J$ stacked documents and $N$ attributes to extract.

\subsection{Fine-tuning and Distillation}
\label{sec:HPD_fine_tuning}
As HPD makes no modification to the LLM architecture or model weights, it can be used directly on pre-trained LLMs. However, there exists a mismatch between the training data and test data.
Firstly, the model can no longer attend to past values when generating a subsequent one, so the attention weights must be redistributed. Secondly, the use of $K_{max}$ as a fixed position ID gap means that values shorter than $K_{max}$ will cause some position IDs to be skipped, which affects the rotary positional embeddings during attention. We find that HPD can be used effectively with LLMs of several sizes at insignificant quality loss, and even quality gains in some scenarios (see \Cref{tab:standard_PAVE}).

To address the train/test mismatch and achieve equal performance to autoregressive decoding, a custom fine-tuning step for HPD is introduced. Given a set of in-domain ground truth labels, we format the training set to mimic the hyper-parallel decoding process by: (a) introducing the same position ID gaps as used in inference, and (b) modifying the causal attention mask to mask value tokens that would not have yet been generated during HPD. In particular, given a set of value labels $\{\bv_{1,1:K_1},\bv_{2,1:K_2},\dots,\bv_{N,1:K_N}\}$, we apply a mask such that each token $v_{n,k}$ cannot attend to tokens $\{v_{n',k'}\, \forall\, n'<n, \, \forall\, k'>k \}$. Similarly, we prevent all attribute tokens from attending to all value tokens, since they will not be available in the first inference step. This mask is applied in addition to the standard causal mask, which prevents tokens from attending to those with a higher position ID. \Cref{fig:HPD_mask} (b) illustrates the attention mask during fine-tuning. The modified mask replicates the attention pattern observed during inference (\Cref{fig:HPD_mask} (a)): when decoding the $k^{th}$ token of a value, the $k+1$ to $K_n$ tokens of all values are missing and thus cannot be attended to.

\begin{table*}[t!]
\centering
\aboverulesep=0ex
\belowrulesep=0ex

\begin{tabular}{@{}llcccccccc@{}}

\toprule
\multirow{3}{*}{Model}      &                          & \multicolumn{2}{c}{\multirow{2}{*}{OA-Mine}} & \multicolumn{2}{c}{\multirow{2}{*}{AE-110K}} & \multicolumn{4}{c}{Amazon Reviews 2023}                                       \\
                            &                          & \multicolumn{2}{c}{}                         & \multicolumn{2}{c}{}                         & \multicolumn{2}{c}{Base model}         & \multicolumn{2}{c}{Fine-tuned model} \\
                            &                          & F1         & \$/1k P.                        & F1         & \$/1k P.                        & F1$_{LLM}$ & \$/1k P.                  & F1$_{LLM}$        & \$/1k P.        \\ \midrule
\multicolumn{2}{l|}{GPT-4(.1) (0-shot)}                          &   0.681     & \multicolumn{1}{c|}{N/A}      & 0.621       & \multicolumn{1}{c|}{N/A}      & \textbf{0.871}       & \multicolumn{1}{c|}{1.14} & N/A                &  N/A           \\ 
\multicolumn{2}{l|}{GPT-4(.1) (10-shot)}                        &     0.822   & \multicolumn{1}{c|}{32.15}      & \textbf{0.875}       & \multicolumn{1}{c|}{17.85}      & N/A       & \multicolumn{1}{c|}{N/A} & N/A                &  N/A           \\ \midrule
\multirow{2}{*}{Qwen3-32B}  & \multicolumn{1}{l|}{AR}  & 0.888       & \multicolumn{1}{c|}{0.688}      & 0.794       & \multicolumn{1}{c|}{0.727}      & 0.791       & \multicolumn{1}{c|}{5.347} & \textbf{0.884}               &  7.745           \\
                            & \multicolumn{1}{l|}{HPD} & 0.878       & \multicolumn{1}{c|}{0.267}      & 0.854       & \multicolumn{1}{c|}{0.270}      & 0.797       & \multicolumn{1}{c|}{0.805} & \textbf{0.884}               &  0.898          \\ \midrule
\multirow{2}{*}{Phi4-14B}   & \multicolumn{1}{l|}{AR}  & 0.880       & \multicolumn{1}{c|}{0.355}      & 0.797       & \multicolumn{1}{c|}{0.350}      & 0.839       & \multicolumn{1}{c|}{2.675} & 0.876               &     3.420        \\
                            & \multicolumn{1}{l|}{HPD} & 0.888       & \multicolumn{1}{c|}{0.095}      & 0.782       & \multicolumn{1}{c|}{0.109}      & 0.814       & \multicolumn{1}{c|}{0.216} & 0.883               &     0.248        \\ \midrule
\multirow{2}{*}{Qwen3-8B}   & \multicolumn{1}{l|}{AR}  & \textbf{0.891}       & \multicolumn{1}{c|}{0.166}      & 0.787       & \multicolumn{1}{c|}{0.163}      & 0.646       & \multicolumn{1}{c|}{1.142} & 0.874               &  1.362           \\
                            & \multicolumn{1}{l|}{HPD} & 0.878       & \multicolumn{1}{c|}{0.069}      & 0.853       & \multicolumn{1}{c|}{0.070}      & 0.751       & \multicolumn{1}{c|}{0.132} & 0.881               &  0.167           \\ \midrule
\multirow{2}{*}{Qwen3-4B}   & \multicolumn{1}{l|}{AR}  & 0.878       & \multicolumn{1}{c|}{0.147}      & 0.795       & \multicolumn{1}{c|}{0.156}      & 0.671       & \multicolumn{1}{c|}{1.120} & 0.861               &     1.200        \\
                            & \multicolumn{1}{l|}{HPD} & 0.876       & \multicolumn{1}{c|}{0.061}      & 0.839       & \multicolumn{1}{c|}{0.063}      & 0.732        & \multicolumn{1}{c|}{0.127}  & 0.870               &   0.133          \\ \midrule
\multirow{2}{*}{Qwen3-1.7B} & \multicolumn{1}{l|}{AR}  & 0.863       & \multicolumn{1}{c|}{0.080}      & 0.780       & \multicolumn{1}{c|}{0.076}      & 0.510       & \multicolumn{1}{c|}{0.887} & 0.861               &     1.110        \\
                            & \multicolumn{1}{l|}{HPD} & 0.849       & \multicolumn{1}{c|}{0.034}      & 0.814       & \multicolumn{1}{c|}{0.035}      & 0.529        & \multicolumn{1}{c|}{0.090}  & 0.870               &   0.095          \\ \bottomrule
\end{tabular}
\caption{F1 scores and cost per 1k products of selected models for AVE on our selected datasets between the autoregressive (AR) baseline and our proposed HPD. Bold numbers indicate the highest F1 score per dataset. LLM-as-a-judge F1 scores are used for Amazon Reviews. 
}
\label{tab:standard_PAVE}
\end{table*}

When no ground truth labels are available, the fine-tuning set can be created as pseudo ground truth from a teacher LLM (Knowledge Distillation), or from the LLM itself using autoregressive generation (self-supervision). In AVE, this process can often be combined with the fine-tuning step that is already performed to align a base LLM to the task, therefore incurring zero additional training cost.

\section{Experimental Setup}
\label{sec:exp_setup}
\paragraph{Datasets}
We evaluate HPD on two standard widely-used AVE benchmarks, OA-Mine \cite{Zhang2022_OA-Mine} and AE110k \cite{xu2019-AE110k}. We select benchmarks containing human-annotated or verified labels over machine-labeled benchmarks such as MAVE that have been found to be more error-prone \citep{zou-etal-2024-implicitave}. We also introduce a third large-scale benchmark based on Amazon Reviews 2023 \citep{hou2024bridging}, which contains both product titles and descriptions. This benchmark is significantly larger in size, more challenging and more costly to extract attributes from. It requires a zero-shot setting as no labels are available. To address this, we employ knowledge distillation, fine-tuning the local LLMs on the zero-shot outputs of GPT-4.1. \Cref{tab:dataset_splits} describes the data split for each dataset, and Appendix \ref{app:datasets} provides the dataset details.
\begin{table}[ht!]
\begin{tabular}{@{}llll@{}}
\toprule
Dataset                                  & Train & Test & Attr./Cat. \\ \midrule
\multicolumn{1}{l|}{OA-Mine}             &  1,452     &  491   &    10.63  \\
\multicolumn{1}{l|}{AE110k}              &   1,568    &  524   &  10.10    \\
\multicolumn{1}{l|}{Amazon Reviews} &  45,000  &  17,905  &   16.00   \\ \bottomrule
\end{tabular}
\caption{The data split breakdown. Attr./Cat. shows the average number of attributes per product category. For Amazon Reviews, pseudo-labels are generated using GPT-4.1 to form the training data.}
\label{tab:dataset_splits}
\vspace{-1em}
\end{table}

\begin{table}
\centering
\begin{tabular}{l|cc}
\toprule
\multicolumn{3}{c}{Amazon Reviews 2023} \\
Model & Prod./s & Speedup \\
\midrule
Qwen3-32B HPD   &  0.539 & 10.78 X \\ \hline
Qwen3-32B + 1.7B Spec.    & 0.104 & 2.08 X \\                        
Qwen3-32B + 4B Spec.  &  0.092 & 1.84 X \\                       
Qwen3-32B + 8B Spec.  &  0.063 & 1.26 X \\ \hline
Qwen3-32B AR &  0.050 & 1.00 X \\
\bottomrule
\end{tabular}
\caption{Inference throughput in product/s of HPD compared to speculative decoding on Amazon Reviews. Qwen3-32B is used as the base model, with differently-sized draft models of the same family. The second column shows speedup relative to standard autoregressive (AR) decoding. Results are shown with no batching and no document stacking.}
\label{tab:spec_dec}
\end{table}

\paragraph{Models}
We use a set of state-of-the-art LLMs across a wide range of model sizes. GPT-4.1 is used as the most performant and expensive baseline\footnote{For OA-Mine and AE-110k, we report gpt-4-0613 results from \citet{Brinkmann2024_ExtractGPT}.}. Since the model is locked behind an API, HPD cannot be used. We instead employ a few-shot setting on OA-Mine and AE110k, which has been shown to provide strong performance \citep{Brinkmann2024_ExtractGPT}. For more cost-efficient inference, we select the Qwen3 family of models in 1.7B, 4B, 8B and 32B sizes \citep{qwen3technicalreport} as a representative range of highly performant decoder-only LLMs.
We also include Phi-4 14B \citep{phi4technicalreport} to represent other model families. Each model is fine-tuned with LoRA. 
Appendix \ref{app:exp_details} contains the full inference and fine-tuning details.

\paragraph{Metrics}
We report cost and throughput as defined in \Cref{sec:problem_setting}. We report the F1 score achieved by each model on OA-Mine and AE110k. For Amazon Reviews, Claude 3.5 Sonnet \citep{claude3.5sonnet} is used as an unbiased judge to evaluate the correctness of extracted values. We report an F1 score derived from the LLM evaluation. The prompt and calculation details are provided in Appendix \ref{app:example_prompt}.

\begin{figure}[t!]
  \centering
  \includegraphics[width=0.95\linewidth]{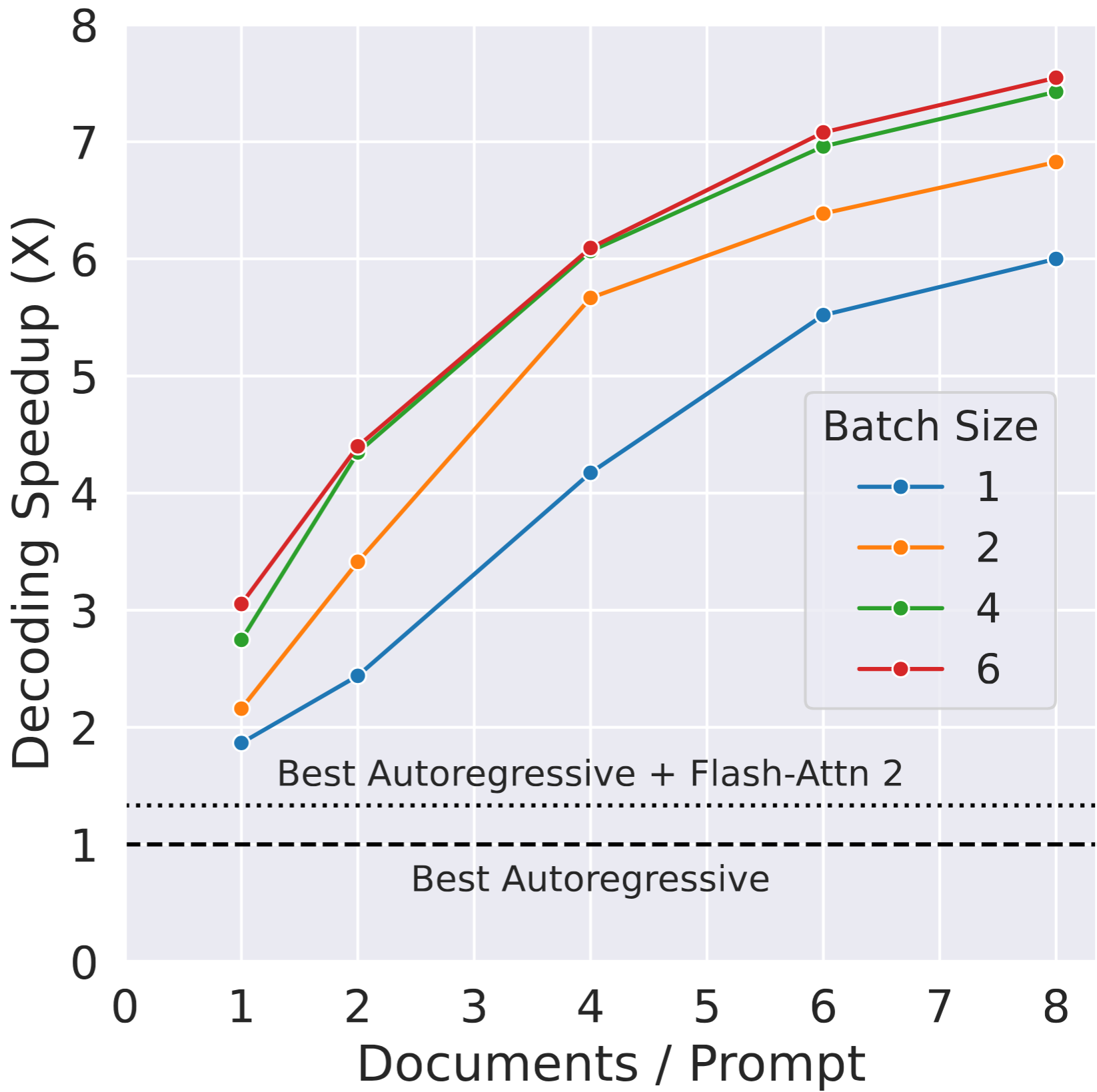}
  \caption{Throughput (products / s) increase from HPD relative to autoregressive decoding on Qwen3-8B for varying documents per prompt and batch size on Amazon Reviews 2023. HPD achieves up to 7.55X speedup. The autoregressive baseline maximizes batch size given VRAM constraints, shown with standard SPDA and Flash-Attention 2.}
  \label{fig:HPD_scaling}
  \vspace{-1em}
\end{figure}

\section{Results} 
\Cref{tab:standard_PAVE} reports the performance and cost of each model with autoregressive decoding (AR) and Hyper-Parallel Decoding (HPD), on all datasets. Results show that HPD reduces the cost of inference by up to \textbf{3.73X}, \textbf{3.21X}, and \textbf{13.79X} on OA-Mine, AE110K and Amazon Reviews for Phi4-14B. The cost reduction is directly proportional to the increase in throughput, given in Appendix \ref{app:throughput} for Amazon Reviews. For the fine-tuned LLMs, this cost reduction comes with no quality penalty: The average F1 score for HPD is on average 0.7\% lower for OA-Mine, but 4.5\% higher for AE110k and 0.7\% higher for Amazon reviews. We make four key observations:

\paragraph{HPD's performance gains depend on the task, but are consistent across a range of model sizes.} The obtained cost reduction on Amazon Reviews is significantly higher than the other datasets. The document context size and number of attributes per category play a significant role in the performance gains of HPD. HPD can parallelize more effectively when there are more attributes to extract, and longer documents reduce the maximum batch size that autoregressive inference can use. Given a fixed dataset, the speedup from HPD is mostly consistent across model sizes ranging from 1.7B to 32B.

\paragraph{HPD is plug-and-play.} When using the base models for Amazon Reviews, we observe that the models using HPD perform on par or better than their autoregressive counterparts on 5/6 of the models tested. This demonstrates that HPD is not limited by requiring fine-tuning.

\paragraph{HPD is preferable over speculative decoding.} We compare HPD directly to speculative decoding on the Amazon Reviews dataset. We select Qwen3-32B as the base model and Qwen3-8B,4B and 1.7B as the draft models, using the Hugging Face Transformers dynamic thresholding implementation. Batching is disabled for this experiment because it is not supported by speculative decoding. \Cref{tab:spec_dec} shows that speculative decoding can increase throughput by up to 2.08X compared to autoregressive decoding on this task. HPD significantly outperforms speculative decoding by achieving a 10.78X speedup. The speedup difference can be attributed to the fact that HPD completely skips the draft phase and can still parallelize inference of the base model, while maintaining output quality.

\paragraph{HPD can linearly scale inference throughput with document stacking.} \Cref{fig:HPD_scaling} illustrates how the relative throughput of Qwen3-8B HPD improves with regards to batch size and number of stacked documents. The autoregressive baseline uses the maximum allowable batch size under VRAM constraints. We observe that throughput scales better with document stacking than batch size. The added parallelism from document stacking linearly increases throughput up to a saturation point, where added parallelism becomes ineffective.
Extrapolating our throughput gains to a corpus of 500M Amazon Reviews-like documents using the cost-effective Qwen3-8B, HPD would achieve a cost saving of \$597,000 over autoregressive decoding, and \$486,000 against GPT-4.1.

\paragraph{Human evaluation results are aligned with the LLM judge preference on Amazon Reviews.}
To validate the LLM judge used for evaluating Amazon Reviews, we conduct a human evaluation experiment on a subset of our results in Appendix \ref{app:human_eval}, showing agreement between the human annotators and the LLM judge. In a blind test, human annotators slightly prefer Qwen3-8B HPD over its autoregressive counterpart as well as GPT-4.1, which aligns with the performance ordering according the the LLM judge.

\section{Conclusion}
Large Language Models excel at solving a wide range of problems, but the sequential nature of autoregressive generation imposes a bottleneck on performance. In some tasks such as AVE, the generated outputs can be broken down into independent components. \hpd leverages this independence to enable LLMs to generate multiple tokens in parallel, without requiring any architectural or model weight modifications. HPD parallelizes generation within each prompt, sharing both memory and computation while synergizing with batched inference, crucial for the offline setting. By manipulating the input prompt, position IDs and attention mask, HPD maintains functionality with key existing methods such as key-value caching, demonstrating a zero-compromise throughput increase and cost reduction on a range of datasets in the product AVE domain. HPD can in theory generalize beyond AVE to other tasks with similar independent structures, which we aim to explore in future work.

\newpage

\section*{Limitations}
\label{sec:limitations}
\paragraph{Models} We perform our experiments primarily using the Qwen3 model family, as well as Phi4. Experiments using other model families and LLMs exceeding 32B parameters are omitted due to computational constraints.
\paragraph{Datasets}In this work, we evaluated the performance of \hpd exclusively on product AVE, although the method does not require product data. We leave for future work to demonstrate the applicability of HPD in other domains and other tasks beyond AVE.
\paragraph{Compatibility}\hpd accelerates inference by reducing the total number of inference steps during generation, and is orthogonal to most other acceleration methods. We have shown compatibility with quantization and knowledge distillation, but have not shown how other methods such as paged attention  and Mixture-of-Experts interact with HPD. One incompatibility of note is Flash-Attention \citep{dao2023_flashattention2}, which assumes a triangular attention mask in its current implementation. However, new implementations such as FlexAttention \citep{dong2024flexattentionprogrammingmodel} now provide similar benefits to FlashAttention with the flexibility of custom attention masks. We provide an implementation of HPD with both FlexAttention and standard scaled dot-product attention (SDPA), although the latter is used in our experiments.
\paragraph{Evaluation} We observed that the labels for AE110k and OA-Mine do not contain all present values, sometimes leading to the LLM being penalized for extracting a correct attribute. For Amazon Reviews 2023, evaluation is performed by an LLM judge due to resource constraints. Although our human evaluation agreed with the LLM judge on a small subset of data, it is not a perfect substitute.

\bibliography{anthology,custom}

\clearpage

\appendix

\section*{Appendix}

\section{The HPD Algorithm} \label{app:Algorithm}

\Cref{alg:HPD} describes the entire \hpd algorithm in pseudo code, leaving out the manipulation of position IDs, which is described in Appendix \ref{app:position_id}. \Cref{fig:app_Figure2} contains is an alternative illustration to \Cref{fig:HPD_flowchart}.

\begin{figure}[ht!] 
\begin{algorithm}[H]
\caption{Hyper-Parallel Decoding}\label{alg:HPD}
\begin{algorithmic}[H]
\State $\bp \gets p(\bx_{1:J}, \ba_{1:N})$
\State $\bs \gets cat(\bp,\ba_{1:N})$
\State $\boldsymbol{values} = empty(J,N,K_{max})$
\State $t \gets 0$
\While{$t < K_{max} \text{ and not all }\boldsymbol{values} \text{ pruned}$}
\If{$t = 0$}
    \State $\boldsymbol{d},\,\boldsymbol{c} \gets \sigma(F(\bs))$ \Comment{Prefill}
\Else
    \State $\boldsymbol{d},\,\boldsymbol{c} \gets \sigma(F(\boldsymbol{d},\,\boldsymbol{c}))$ \Comment{Autoregressive}
\EndIf
\State $\boldsymbol{d} \gets prune(\boldsymbol{d})$
\For{$j \in range(J)$}
\For{$i \in range(N)$}
\If{$\boldsymbol{values}[j,i] \text{ not pruned}$}
\State $\boldsymbol{values}[j,i,t] \gets \boldsymbol{d}[jN+i]$
\EndIf
\EndFor
\EndFor
\State $k \gets k+1$
\EndWhile
\end{algorithmic}
\end{algorithm}
\vspace{-1em}
\caption{The basic \hpd algorithm. Given a prompt $\bp$ to extract $N$ values for attributes $\ba_{1:N}$ from $J$ products $\bx_{1:J}$, perform up to $K_{max}$ inference steps, where $K_{max}$ is the maximum value length. Each step generates $j\times N$ tokens $\boldsymbol{d}$, and updates the key-value cache $\boldsymbol{c}$. Step $t$ generates the $t^{th}$ token off all values in parallel, and completed values are pruned. The indexing used assumes no pruning for illustrative purposes.}
\label{fig:HPD_alg}
\end{figure}

\subsection{Position ID calculation} \label{app:position_id}
We begin by assigning the input $\bs=cat(\bp,\ba_{1:N})$ containing the prompt and skeleton output with standard initial position IDs $pos_{id} =[0,1,2,\dots,|s|]$. Let $\text{start}(\ba_i),\text{end}(\ba_i)$ be the start and end token index of attribute $\ba_i$ in $\bs$. We then insert position ID spacing of $K_{max}$ between each attribute according to \Cref{alg:pos_id}. During inference, the $k^{th}$ token of the $n^{th}$ attribute $v_{n,k}$ is assigned position ID: $pos_{id}[\text{end}(\ba_n)] + k$.

\begin{figure}[ht!] 
\begin{algorithm}[H]
\caption{Position ID assignment}\label{alg:pos_id}
\begin{algorithmic}[H]
\State $\bs \gets cat(\bp,\ba_{1:N})$
\State $pos_{id} = range(|\bs|)$
\State $\text{offset} \gets 0$
\For{$i \in range(N)$}
\State $pos_{id}[\text{start}(\ba_i)\text{:}\text{end}(\ba_i)+1] \pluseq \text{offset}$
\State $\text{offset} \gets \text{offset} + K_{max}$
\EndFor
\end{algorithmic}
\end{algorithm}
\vspace{-1em}
\caption{Position ID assignment algorithm for the model input $\bs$. After assigning initial position IDs, a space of $K_{max}$ is inserted in between each attribute for the decoded values to be inserted.}
\label{fig:pos_id}
\end{figure}

\begin{figure*}[ht!]
  \centering
  \includegraphics[width=0.85\textwidth]{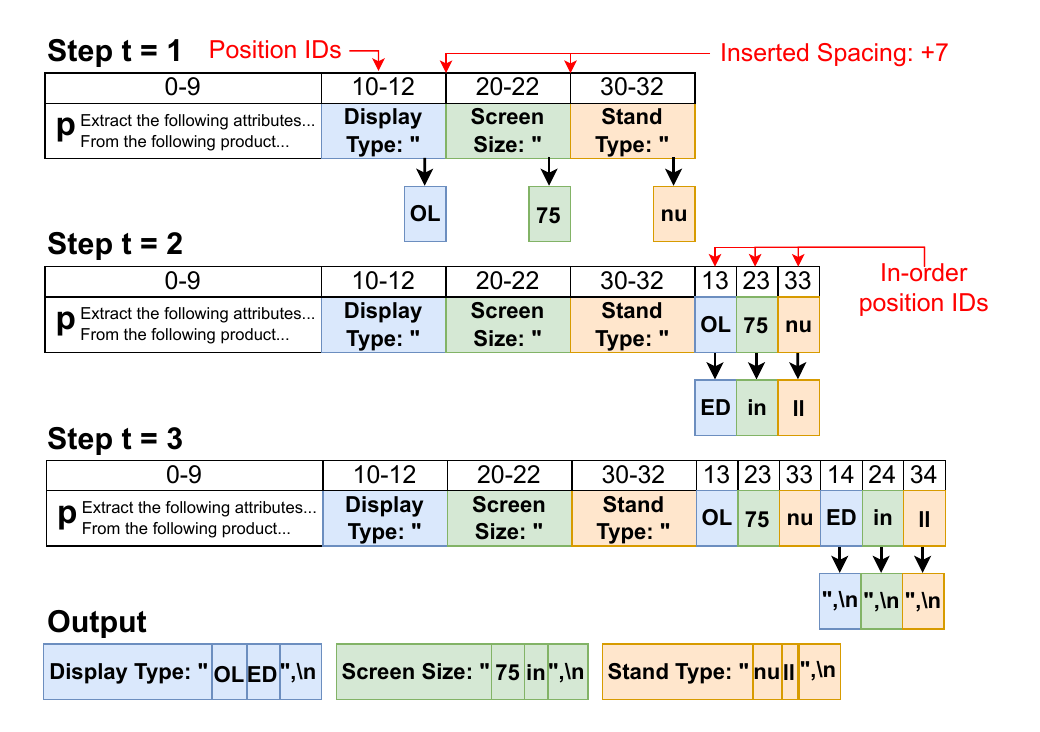}
  \caption{Alternative block diagram for the HPD process: At step $t=1$, the first token of each value is decoded in parallel using the next token prediction at each attribute end position. The position IDs shown above each token have an inserted gap between attributes, which is gradually filled by the generated value tokens at each step. New tokens are always appended to the sequence in memory which places them out of logical order, but the position IDs define the attention mask and positional embeddings during attention. In this example, three values of three tokens each are decoded in parallel using only three inference steps.}
  \label{fig:app_Figure2}
\end{figure*}

\section{Additional Experimental Details}
\label{app:exp_details}
\subsection{Dataset Details} \label{app:datasets}
\paragraph{OA-Mine \cite{Zhang2022_OA-Mine}:} We use the human-annotated subset of OA-Mine, containing 9,811 attribute-value pairs for 1,943 products across 10 product types from Amazon.com. The data consists of only product titles. We follow the same (large) train/test split used by \citep{Brinkmann2024_ExtractGPT}.

\paragraph{AE110k \cite{xu2019-AE110k}} AE110k contains 39,505 product titles from AliExpress Sports \& Entertainment, with the label values obtained directly from their structured product catalog. There are a total of 2,045 unique attributes and 10,977 unique values. We follow the same (large) train/test split used by \citep{Brinkmann2024_ExtractGPT}.

Although these two AVE datasets contain high quality labels useful for correctness evaluation, they differ substantially from the large-scale e-commerce datasets where LLM inference cost is a primary concern. Firstly, product information is not only contained in product titles, but also in product descriptions. The inclusion of product description not only makes the task more challenging, but also more costly due to the longer product context and number of attributes present. Secondly, products from large e-commerce services have a wider range of product categories than the subsets selected in OA-Mine and AE110k. Therefore, acquiring ground truth labels is often costly or infeasible, requiring a zero-shot setting. We therefore craft a new benchmark using the open source product data from Amazon Reviews 2023 \citep{hou2024bridging}:

\paragraph{Amazon Reviews:} We collect 900k products across 31 product categories. We use Claude 3.7 Sonnet \citep{claude3.7sonnet} to define the 16 most important attributes for each product category, resulting in 267 unique attributes in total. Due to the absence of labeled training data, we generate zero-shot predictions using GPT-4.1 with batched API \citep{gpt4.1} on 45k uniformly sampled products, using their titles, descriptions, and bullet points as input. We then use these predictions as pseudo-labels to fine-tune small language models (SLMs) through knowledge distillation. Evaluation is conducted on a separate sample of 18k products. The relevant prompts are provided in Appendix \ref{app:example_prompt}.

\subsection{Fine-tuning and Inference Details}
All experiments are run on the Hugging Face Transformers framework with Accelerate. We fine-tune the local models for 5 epochs on OA-Mine and AE110k. A separate fine-tuning is performed for standard autoregressive inference and HPD using the custom alignment process described in Section \ref{sec:HPD_fine_tuning}. DeepSpeed Zero3 and LoRA are used to efficiently fine-tune even the Qwen3-32B model. On Amazon Reviews, we apply knowledge distillation for 1 epoch. The learning rate is tuned as a hyperparameter for the lowest validation loss. For Qwen3-4B and 1.7B, we use full-parameter finetuning.

At inference, we stack 6 documents per prompts, and use a maximum value length $K_{max}=30$. 4-bit quantization is used for Qwen3-32B with NF4 using BitsandBytes. For smaller models, we use bf16 inference as we find quantization to not improve inference speed given the available VRAM. The batch size for each model is adjusted to max out the VRAM utilization and ensure fair comparison between the larger stacked prompts in HPD and standard autoregressive inference. We use Flash-Attention-2 for autoregressive inference, but revert to SPDA for HPD since Flash-Attention does not support modified attention masks by default.

\subsection{Resources}
We select the highly performant Amazon EC2 g6e.48xlarge server for local fine-tuning and inference, using 8-way data parallelism on Nvidia L40S 48GB GPUs. The cost/product of the local models is derived from the time required to process all products on the test set on this instance and the on-demand rental cost of \$30.13/h as of July 2025. We select this instance because it is publicly available and representative of the type of server that would be used for efficiently processing millions of products for AVE.For API based LLMs, we define cost as the average API credit cost/product (\$/1k products) as of July 2025. This cost already includes a discount for prefix caching.

\section{Throughput Measurements on Amazon Reviews} \label{app:throughput}
Table \ref{tab:AR_throughput} shows the throughput in products processed per second for the local fine-tuned models on Amazon Reviews. We observe a relatively constant throughput increase of 10X across model sizes.
\begin{table}
\centering
\begin{tabular}{ll|cc}
\toprule
& & \multicolumn{2}{c}{Amazon Reviews 2023}\\
Model & & LLM-F1 & Prod./s \\
\midrule
\multirow{2}{*}{Qwen3-32B} & AR  & 0.884 & 0.14 \\
                           & HPD & 0.884 & 1.17 \\
\hline
\multirow{2}{*}{Phi4-14B} & AR   & 0.876 & 0.31 \\
                           & HPD  & 0.883 & 4.23 \\
\hline
\multirow{2}{*}{Qwen3-8B} & AR  &  0.874 & 0.77 \\
                           & HPD  & 0.881 & 6.28 \\
\hline
\multirow{2}{*}{Qwen3-4B} & AR   & 0.861 & 0.87 \\
                           & HPD & 0.870 & 7.89 \\
\hline
\multirow{2}{*}{Qwen3-1.7B} & AR & 0.861 & 0.94 \\
                           & HPD & 0.870 & 10.99 \\
\bottomrule
\end{tabular}
\caption{LLM-as-a-judge F1 scores and throughput (products/s) of selected fine-tuned local models for Amazon Reviews. We compare the autoregressive (AR) and hyper-parallel (HPD) performance for each local model.}
\label{tab:AR_throughput}
\end{table}

\section{Human Evaluation on Amazon Reviews 2023} \label{app:human_eval}
To further validate the LLM-as-a-judge F1 scores on Amazon Review 2023, we conducted a blind human evaluation on 563 examples to assess the quality of LLM evaluation. Four expert annotators performed pairwise comparisons between outputs from GPT-4.1, Qwen-8B (AR), and Qwen3-8B (HPD), deciding which model's extractions are most faithful to the product context. For each comparison, annotators examined the original review text alongside attribute-value pairs from two randomly selected models, then judged whether one output was superior (win), inferior (loss), or comparable (tie) to the other. 

\begin{figure}[ht!]
  \centering
  \includegraphics[width=\linewidth]{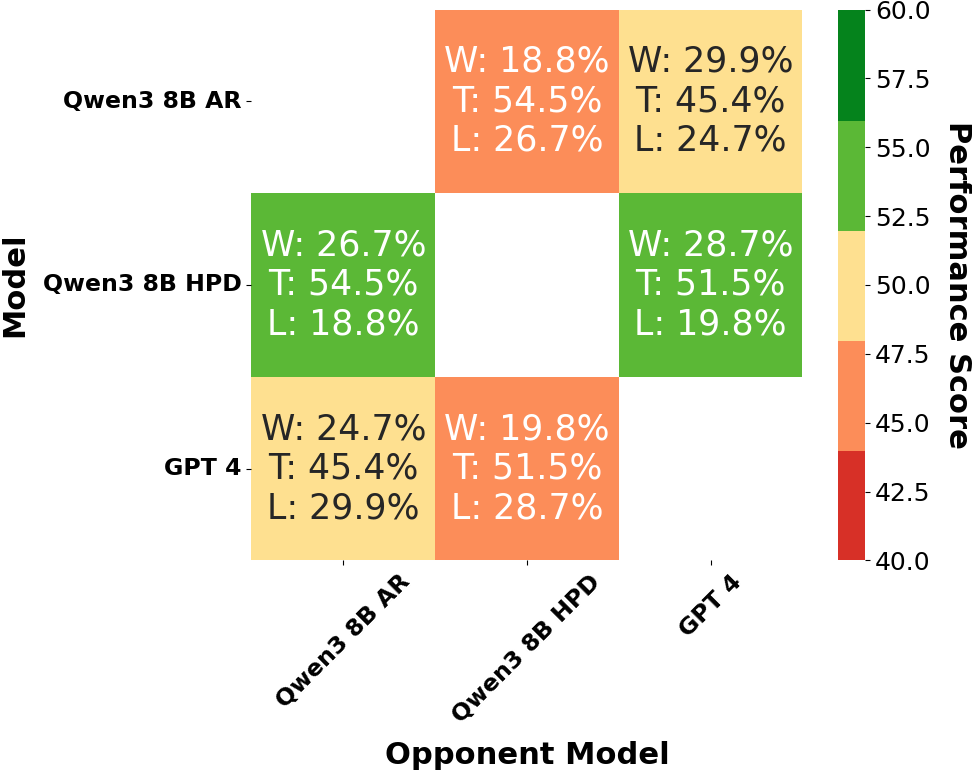}
  \caption{Amazon Reviews 2023 dataset: Matrix of win (W), tie (T) and loss (L) rate of GPT-4.1, standard Qwen3-8B (AR), and Qwen3-8B with HPD. Wins, ties and losses are determined by human annotators randomly comparing 2/3 model outputs for one product, and judging which set of values are most faithful to the product context. Performance score is calculated as (W+T)/(T+L) normalized to 0-100.}
  \label{fig:human_eval}
\end{figure}

Figure \ref{fig:human_eval} presents the comparison matrix with win/tie/loss rates across the three models. Qwen3-8B (HPD) achieves higher win rate against GPT-4.1 (28.7\% vs. 19.8\%), surpassing even the teacher model in human preference. When comparing the two student models directly, Qwen3-8B (HPD) also outperforms its AR counterpart—winning 26.7\% of match-ups while losing only 18.8\% and with 54.5\% ties, demonstrating that our approach not only achieves comparable output quality against the teacher model and the autoregressive approach, but also drastically reduces the inference time.

\section{Example Prompts and Outputs}
\label{app:example_prompt}
Figures \ref{code:PAVE_prompt}, \ref{code:attr_def_prompt}, \ref{code:eval_prompt} contain the prompts for extracting attribute values, defining the product category attributes and evaluating the quality of extracted values for Amazon Reviews. Using the defined classification categories defined in Figure \ref{code:eval_prompt}, we define $C=$ "correct", $CN=$ "correct null", $I=$ "incorrect", $M=$ "missing" and $H=$ "hallucination". The LLM F1 score is calculated as:
\begin{align*}
    \text{LLM-P} &= \frac{C + CN}{C+CN+H+I}\\
    \text{LLM-R} &= \frac{C+CN}{C+CN+M+I}\\
    \text{LLM-F1} &= \frac{2\times \text{LLM-P} \times \text{LLM-R}}{\text{LLM-P}+\text{LLM-R}}
\end{align*}

\begin{figure*}[ht] 
  \centering
  \begin{lstlisting}[style=promptstyle]
You are an expert at extracting product details.

Consider the following product attributes:

{attributes}

Analyze each input <product> element below and extract all of the attributes specified above in the requested standard format.

Output a JSON dict with a key for each input product ID, and a nested dict with a key for each attribute (use the attribute name) and the extracted value.
Escape any output double quotes characters (") to ensure a valid output JSON.

The values in the standard format: 
    - must preserve names of characters, themes, brands, patterns, flavors if present. 
    - for Product Type, Product Intent, Sustainability attributes: must preserve details like method names and types without aggregating. 
    - must be 4-5 words or less.
    - must retain all the different numbers and their standard units (e.g., counts, packs), without multiplying them together.
    - must not use placeholders like '-' or 'null' for concise values.
    - must simplify words to base noun forms without plurals, uppercase, adjective or adverbial forms.    
    - must write null if the value of an attribute is not explicitly decribed in the product information.

Put the JSON list in <result> tags.

Input products:

{products}
  \end{lstlisting}
  \caption{Amazon Reviews Prompts}
  \label{code:PAVE_prompt}
\end{figure*}

\begin{figure*}[ht] 
  \centering
  \begin{lstlisting}[style=promptstyle]
You are an expert at comparing products to determine if they are equivalent in terms of function, specification, form, design, material, quantity, quality, brand value.

Analyze the <product> elements below, and output a list of price-sensitive attributes that experienced customers and product category experts would consider when comparing these products to see if they are exact equivalents, very similar, or incompatible.

For each attribute write a brief description which lists 3-6 typical values as examples.
Make sure each attribute name is not more than 4 words.
Do not include attributes related to price.
For each attribute output the data type of its value as numerical or categorical.

Provide up to 16 attributes, including fine-grained ones.

Only output a list of XML <attribute> elements with numeric id (as element attribute), product attribute <name>, <description> and <datatype>.

Products:

{products}
  \end{lstlisting}
  \caption{Attribute definition prompt for Claude 3.7 Sonnet}
  \label{code:attr_def_prompt}
\end{figure*}

\begin{figure*}[ht]
  \centering
  \begin{lstlisting}[style=promptstyle]
You are an expert at evaluating product details.

Consider the following product attributes:

{attributes}

An existing system has attempted to extract the value of each attribute from the product information below. Analyze the input context and extracted values to determine if the extracted values are correct.

Classify each attribute into 1 of 5 types:
- "correct": The input text contains sufficient information to extract the attribute value beyond resonable doubt. The attribute value extracted is correct and complete. Words that convey the same meaning are allowed. 
- "incorrect": The input text contains sufficient information to extract the attribute value beyond resonable doubt. The attribute value extracted is incorrect or incomplete. Do no include "null" values in this category.
- "missing": The input text contains sufficient information to extract the attribute value beyond resonable doubt. The attribute value extracted is "null" or missing.
- "correct null": The value cannot be directly inferred from the input context. The attribute value extracted is "null" or similar. Values indicating the absense of an attribute ("no", "none") should be "correct null".
- "hallucination": The value cannot be directly inferred from the input context. An attribute value was extracted which is not indicating the absence of a value.

Use the following explanation structure for each attribute:
<attribute_name>: The value "<value>" is <classification> because <brief explanation>.

After the explanation, output a JSON dict with a key for each attribute (use the attribute name) and one of the evaluation results as value (correct, incorrect, missing, correct null, hallucination). Wrap the JSON around <result></result> tags. Do not provide explanations.

Input product:

{product}

Extracted Values:

{values}

  \end{lstlisting}
  \caption{Evaluation prompt used for Claude 3.5 Sonnet}
  \label{code:eval_prompt}
\end{figure*}

\end{document}